\documentclass[letterpaper, 10pt, conference]{./ieeeconf}   % Use this line for a4
\IEEEoverridecommandlockouts                              % This command is only
\usepackage[]{algorithmic}
\usepackage[]{algorithm}

\usepackage{graphics, amsfonts, amsmath, amssymb, enumerate, ae, balance}
\usepackage{graphicx}
\usepackage{color}
\usepackage{url}
\usepackage{palatino}%
\makeatletter
\let\NAT@parse\undefined
\makeatother
\usepackage[square, numbers]{natbib}

\renewcommand{\baselinestretch}{1}

\title{\LARGE \bf
	A Fast Randomized Method to Find Homotopy Classes \\ 
	for Socially-Aware Navigation}

\author{
Luigi Palmieri \and Andrey Rudenko \and Kai O. Arras
\thanks{L. Palmieri, A. Rudenko, K.O. Arras are with the Social Robotics Lab, Dept. of Computer Science,
         University of Freiburg, Germany.
         {\{palmieri,arras\}@cs.uni-freiburg.de},~{andrey.rudenko@saturn.uni-freiburg.de}}.%
}

\DeclareMathOperator{\Persons}{\mathcal{P}}

\DeclareMathOperator{\idirection}{\widehat{\boldsymbol{e}}}
\DeclareMathOperator{\fdirection}{\boldsymbol{n}}

\begin{document}

\maketitle

%\thispagestyle{empty}
%\pagestyle{empty}

% ================================================================== %
\begin{abstract}
%In social-aware navigation with scenarios full of pedestrians and dynamic obstacles, having a set of paths from where to choose a valid one in case of the appearance of unexpected obstacles, is a less expensive approach than solving from scratch the motion planning problem for each change of the environment. 
We introduce and show preliminary results of a fast randomized method that finds a set of $K$ paths lying in distinct homotopy classes. We frame the path planning task as a graph search problem, where the navigation graph is based on a Voronoi diagram. The search is biased by a cost function derived from the social force model that is used to generate and select the paths.  We compare our method to Yen's algorithm, and empirically show that our approach is faster to find a subset of homotopy classes. Furthermore our approach computes a set of more diverse paths with respect to the baseline while obtaining a negligible loss in path quality.
%We prove that our approach finds all possible paths in a undirected weighted graph with probability greater than zero.

\end{abstract}

% ================================================================== %
\section{Introduction}
\label{sec:introduction}
%% --
In socially-aware navigation, like in the case where a robot assists, informs and guides passengers in large and
busy airports \cite{triebel15spencer}, the motion planner deals with non-static scenarios where crowds of pedestrians and dynamic obstacles regularly invalidate paths generated by using standard approaches (e.g. A* \cite{nilsson1971problem}, RRT \cite{lavalleICRA99}, PRM \cite{kavraki1996probabilistic}). Each time a path is invalidated, a new motion planning problem has to be solved or in the case of replanning algorithm like D* \cite{koenig2002d} a repairing phase should recompute a new path by first updating costs over a grid map.
 Having a set of $K$ precomputed distinct paths, that may be checked for validity in case of the appearance of unexpected obstacles, is a more reasonable approach than solving from scratch the motion planning problem or to replan for each environment's change. Moreover a more rational approach would be to generate $K$ distinct paths from $K$ different homotopy classes. A homotopy class is defined by the set of paths with the same start and goal points which can be continuously deformed into one another without intersecting obstacles.
%We introduce and show preliminary results of a fast randomized method that finds a set of $K$ paths, according to a social cost, which lie in distinct homotopy classes.

Different approaches have been already introduced to generate a set of paths belonging to different homotopy classes.
%\section{Related Work}
Demeyen and Buro \cite{demyen2006efficient} introduce a method for efficient path planning that searches on a graph built using constrained Delaunay triangulations. The obstacles are described via polygonal representation. The paths, found in the graph, represent different homotopy classes.
%\item 
Eriksson \emph{et al} \citep{erikssonSODA15} find $K$ homotopy classes solving the $K$ shortest paths problem in a two-dimensional environment with polygonal obstacles. They introduce the $K{th}$ shortest path map: a map of the entire free workspace, partitioned into equivalence class regions such that the $k_{th}$ shortest path from a start vertex $s$ to any point in a single region has the same structure. Moreover they introduce a simple visibility-based algorithm, based on Yen's algorithm \cite{yen1971finding}, for computing the $K$ shortest paths between two fixed points.
% grid based obstacles
%\item  
Bhattacharya \emph{et al.} \citep{bhattacharya2010search} propose a method to find different homotopy classes based on A* search over an augmented graph. The graph represents the topological information via the H-signature, a complex analysis value that characterizes a homotopy class, the graph may contain multiple paths to the goal within the same homotopy class.
%\item TODO check if we use directed graph!
Kuderer \emph{et al} in \citep{kuderer2014online} select \emph{K} best homotopy classes by generating \emph{K} shortest paths using Katoh's algorithm \citep{katoh1982efficient}. During navigation the paths feed an optimization algorithm used to generate homotopically distinct kinodynamic trajectories. Among those the best one is selected for the navigation. They show the method is one order of magnitude faster than \citep{bhattacharya2010search}.
\begin{figure}[t!]
  \centering
  \includegraphics[width=0.990\columnwidth]{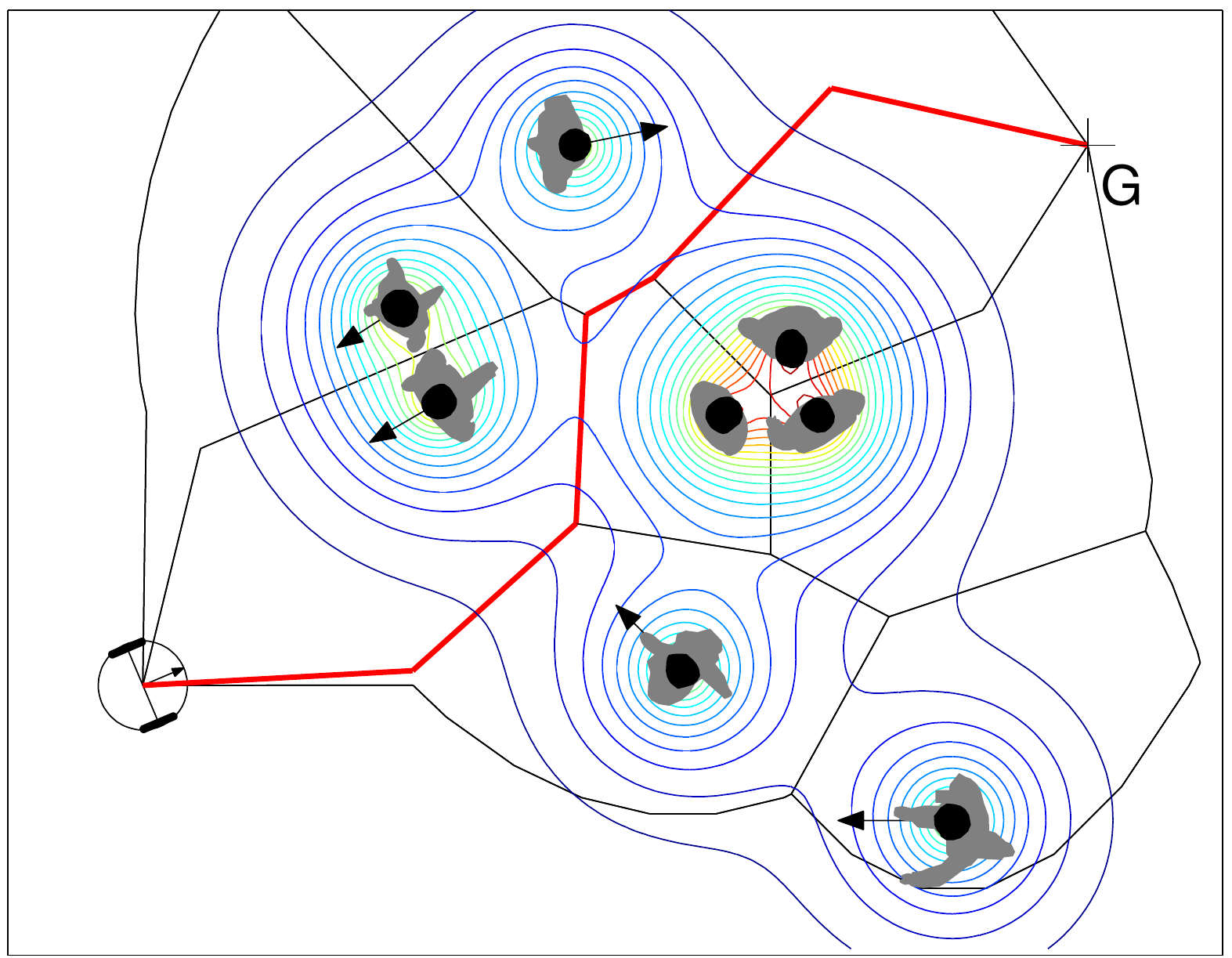}
  \caption{An example path selected from the $K$ best homotopy classes in the Voronoi diagram. The robot is enclosed in the \textbf{black circle}, in \textbf{red} the path selected to reach the goal position \textbf{G}. The \textbf{black} Voronoi diagram describes the possible ways to go through a crowd by implicitly encoding different homotopy classes.}
  \label{fig:robotamongpeople}
\end{figure}
%\item 
Vela \emph{et al} \cite{vela2010} detail a decision support tool to aid air traffic controllers and managers in re-routing traffic: they generate a set of homotopy classes for a given pair of start and goal poses, by computing the $K$ shortest paths via the Dijkstra algorithm on a Voronoi graph.%: the Voronoi diagram of a set of points is dual to its Delaunay triangulation.
   ~The latter is generated from a map that encodes weather conditions. A final step optimize the set of $K$-shortest paths with respect to path length and weather avoidance.
%
% Roadmap construction
%\item 
Voss \emph{et al} \cite{vossICRA2015} introduce an algorithm that seeks to find a set of diverse, short paths through a roadmap graph. The algorithm finds diverse homotopy classes by finding diverse shortest paths avoiding a collection of balls imposed on the graph as simulated obstacles. The authors compare their approach to the Eppstein algorithm \cite{eppsteinSIAM98} that finds the optimal set of ${K}$ shortest paths with loops and show that, with tolerable loss in shortness, they produce equally diverse path sets orders of magnitude more quickly.
% Homotopy A* and RRT
%\item 
Hernandez \emph{et al} \cite{hernandezRAS2015} propose and compare three different path planning algorithms that exploit the set of homotopy classes generated for any 2D workspace: Homotopic A*, Homotopic RRT and Homotopic Bug. Their method to generate homotopy classes modifies the one introduced by Jenkins \cite{jenkinsMASTER1991}: it first builds a reference frame in the workspace which is used to identify the homotopy classes and afterwards it builds a topological graph which allows an easy and systematic computation of homotopy classes. The homotopy classes are sorted according to a lower bound heuristic estimator, then they are used to guide and to constrain topologically the path search.

Instead of solving the $K$ shortest paths problem with deterministic graph search algorithms %methods based on graph search and focusing on finding the $K$ shortest paths 
like in \cite{bhattacharya2010search,kuderer2014online, vela2010}, we introduce and show preliminary results of a fast randomized method based on random walks that finds a set of $K$ homotopically distinct paths, according to a social cost. As in \cite{kuderer2014online} we build a navigation graph from the Voronoi diagram. Each path found in the Voronoi diagram represents a distinct homotopy class. % which lie in distinct homotopy classes.% To find $K$ homotopy classes faster than search based approaches (like the one used in \cite{bhattacharya2010search,kuderer2014online, vela2010}) we adopt a randomized approach. 
 ~Differently from \cite{vossICRA2015, kuderer2014online} we compare our method to Yen's algorithm, a fast algorithm that finds loopless paths. In \cite{vossICRA2015} the authors compare their approach to Eppstein's algorithm which finds paths with loops therefore having a lower chance to find a more diverse set of paths. In \cite{kuderer2014online} the authors use Katoh's algorithm, however, it was shown by Brander and Sinclair \cite{brander1996comparative} that for small size graphs and paths of small number of vertices, like in the case we consider, Yen's is faster than Katoh's. Furthermore in our approach, instead of using a polygonal representation of the obstacles as in \citep{demyen2006efficient,erikssonSODA15}, we use occupancy grids where obstacles are represented by blocked cells, therefore permitting an easier integration with existing mapping frameworks. We assume in our work that the pedestrians' poses are given by a people (or group) tracker \cite{linderFUSION14,luberIJRR11}.

 %%-- Contribution of the paper\\
The contribution of our paper is as follows:
\begin{itemize}
\item we introduce a fast, and easy to implement, randomized approach to find a set of $K$ paths belonging to $K$ different homotopy classes.%We show that the algorithm finds with a probability greater than zero all the paths in the graph;

\item we perform an extensive evaluation and compare our method to Yen's algorithm. Our approach is faster than Yen's to find a subset of homotopy classes in a Voronoi diagram;

\item moreover our approach generates more diverse paths than Yen's (the robot has a more diverse set from where to choose the path to follow), while obtaining a negligible loss in path quality.
\end{itemize}

The paper is structured as follows: we detail our new approach and its analysis in Section \ref{sec:approach}. We present the experiments in Section
 \ref{sec:experiments} and discuss the results in Section \ref{sec:results}. Section \ref{sec:conclusions} concludes the paper.

% ================================================================== %
\section{Our Approach}
\label{sec:approach}

 %Fast method to generate a set of K random paths, many of which are \emph{good}. Nevertheless, if exact set of K best paths is required,  then nothing beats direct K shortest paths search algorithm, like Yen's.

%We follow the Random Walk approach: start from the initial robot position. In each node follow an edge at random. Probability of following an edge is inversely proportional to the length of the edge. Delete each node after leaving it. Stop when there are no outgoing edges in current node. 

We introduce a probabilistic approach to find paths belonging to different homotopy classes for socially-aware robot motion planning. Our approach is complete, it can find all the possible paths, namely all the homotopy classes implicitly encoded in the navigation graph built from a Voronoi diagram that describes the scenario. Having a set of possible paths, we choose the best one according to a cost that considers social interactions between humans.

\subsection{Navigation Graph}
\label{subsec:navigationgraph}
\begin{figure*}[!]
  \centering
  \includegraphics[width=0.65\columnwidth]{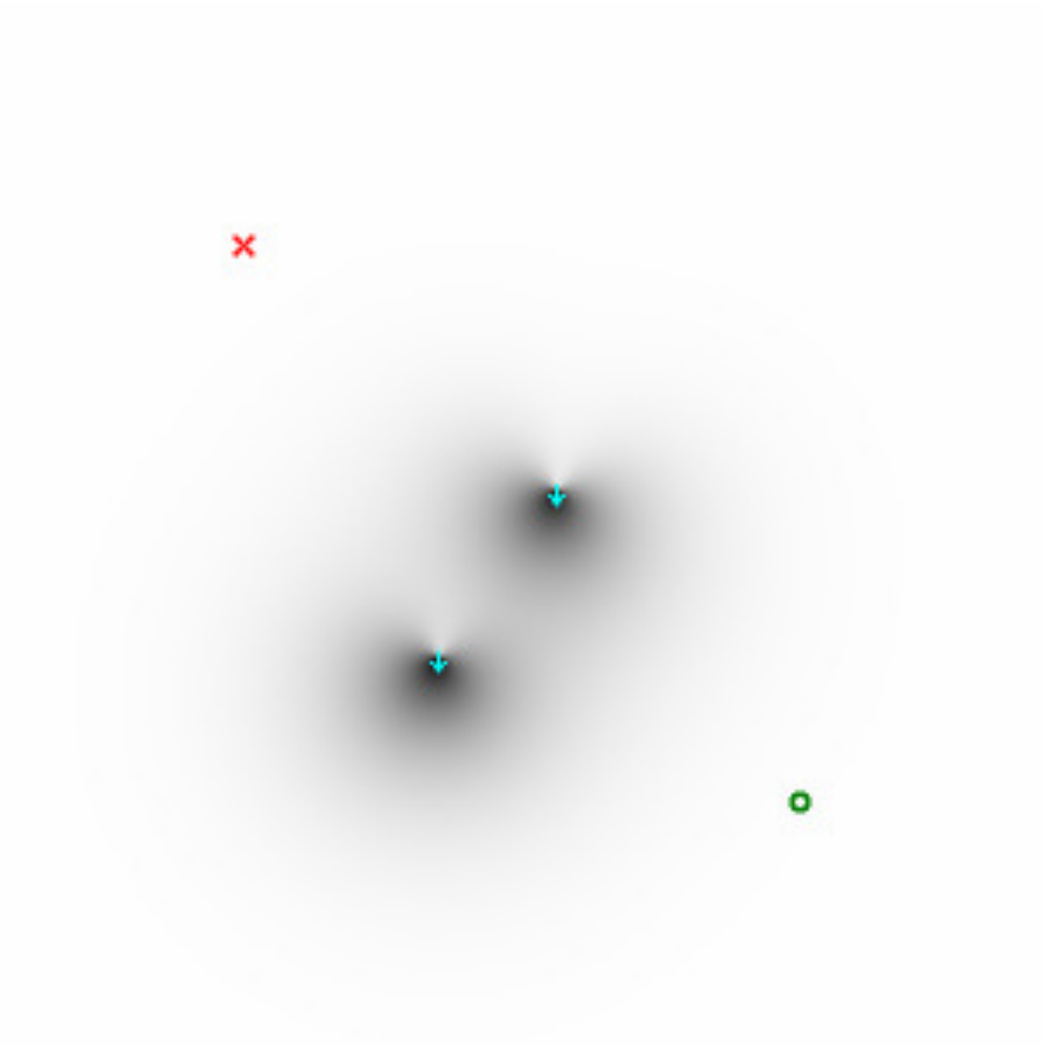}
  \includegraphics[width=0.65\columnwidth]{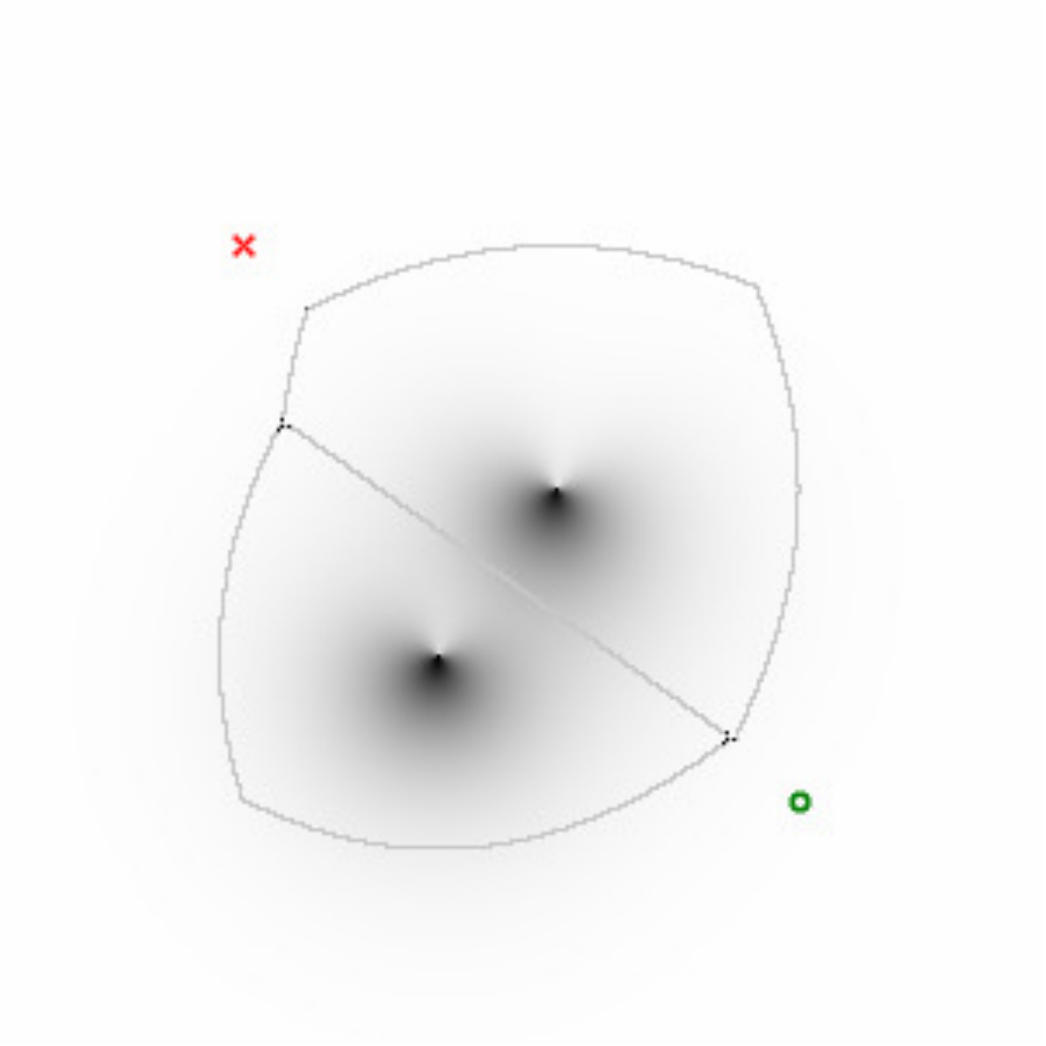}
    \includegraphics[width=0.65\columnwidth]{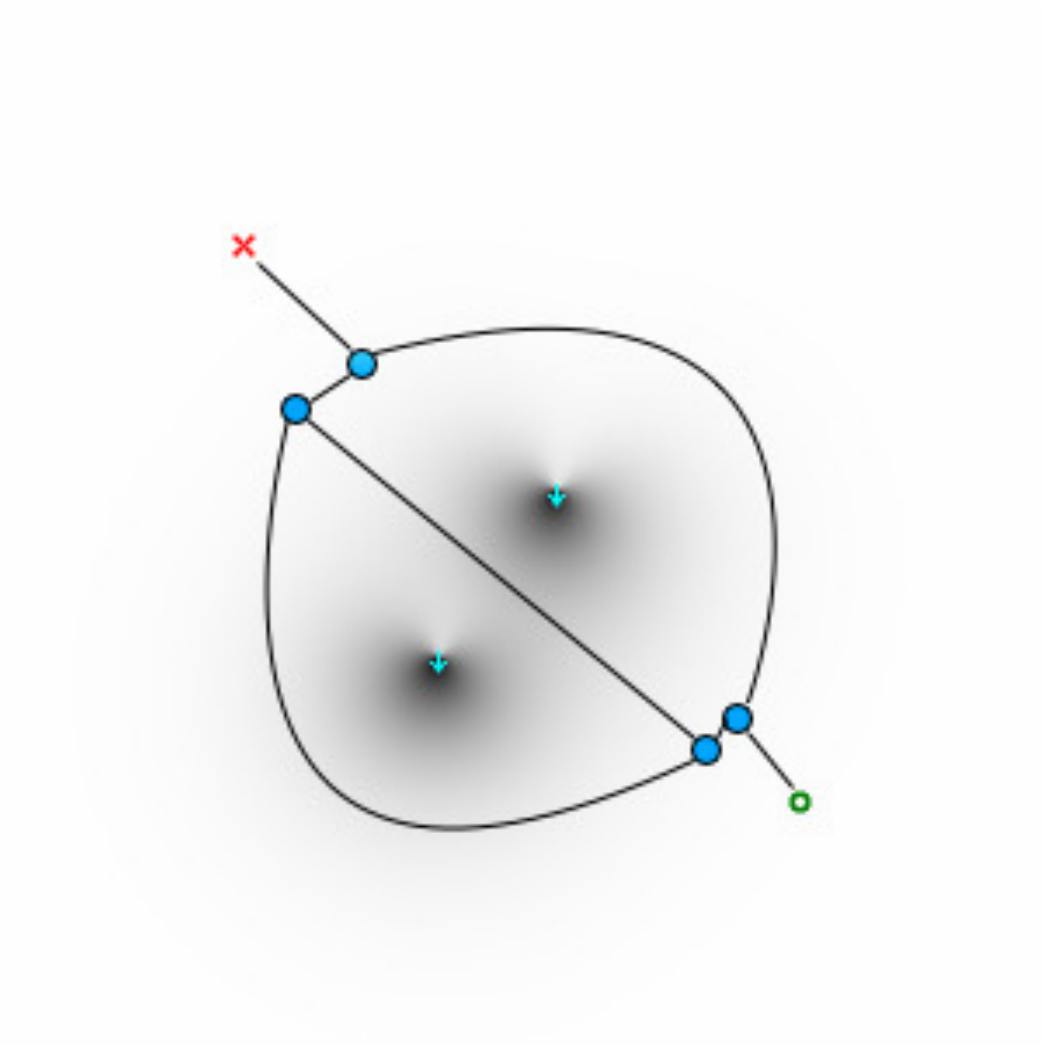}
    \\

  \caption{In all the figures the \textbf{red} cross is the robot position, the \textbf{green} circle is the goal. \textbf{Left:} pedestrians' poses, marked with \textbf{blue} arrows, are provided by the people tracker, a social force field is generated over the two pedestrians. \textbf{Middle:} a Voronoi diagram is built considering the pedestrian positions as obstacles. The Voronoi diagram points are displayed in \textbf{darker grey}. \textbf{Right:} navigation graph built from the Voronoi diagram. The robot and goal nodes are mapped to their closest Voronoi points. The navigation graph \textbf{vertices} are depicted in \textbf{blue}, its edges in \textbf{black} .}
  \label{fig:graphbuilding} 
\end{figure*}

Our method could be used with discrete information received from a people tracker, however, it is equally well suited for use on general occupancy grids. In our case the input scenario is given as a collection of discrete people poses, $(x,y,\theta)$, in the 2D workspace, see Fig.\ref{fig:graphbuilding} (left). To frame the motion planning task as a graph search problem, we build the navigation graph \emph{G} of the scenario from the Voronoi diagram \emph{VD}, generated considering as obstacles also the people poses, see Fig.\ref{fig:graphbuilding} (middle).  We create two additional vertices for the initial robot position and goal position, and connect them to the closest point of the Voronoi diagram, see Fig.\ref{fig:graphbuilding} (right). In the navigation graph, built in this way, different paths from the initial robot position to the goal position belong to different homotopy classes.

~The graph \emph{G(V,E)} consists of a set of nodes (or vertices) \emph{V} and a set of edges \emph{E}. In this work, \emph{N} is the number of nodes in the graph and \emph{M} the number of edges. We associate to each edge ${e_{ij}}$, connecting the node ${v_j}$ to its neighbor ${v_i}$, an attribute or cost $c_{ij}$. $E(v_j)$ denotes the set of incoming and outgoing edges of $v_j$.

 We compute the set of homotopy classes by running our random walk based algorithm on \emph{G}.~% which is generated from the Voronoi diagram \emph{VD} built on a grid using the technique described in \citep{kuderer2014online}. 
A walk \emph{w} of length $k-1$ in a graph is a sequence of nodes $v_1, v_2, \ldots ,v_k$, where each pair of nodes is connected by an edge, $(v_{i-1}, v_i) \in E$ for $1 < i \le k$ . 
The adjacency matrix $A$ of $G$ expresses the topology of the graph and is defined as

\[ [A]_{ij} =
  \begin{cases}
    1      & \quad \text{if } (v_{i}, v_j) \in E, i \neq j\\
    0  & \quad otherwise\\
  \end{cases}
\]
\emph{Walks} are usually referred to as \emph{paths}.%Different type of paths on the graph exist: \emph{not simple} or \emph{with loop} , \emph{unique}, \emph{simple} or \emph{loopless} paths. %In this work we are interested in \emph{simple} path, a walk without repetitions of vertices.

\subsection{Randomized Homotopy Classes Finder}
To find different homotopy classes we introduce the Randomized Homotopy Classes Finder (\textbf{RHCF}), detailed in Alg.\ref{alg:RHCF}. We iteratively run the random walk algorithm, see Alg.\ref{alg:RandomWalk}, on the weighted undirected graph until $K$ distinct paths are found. 

Given the weighted graph $G$, we start at the initial node ${v_s}$, the one where the robot position is mapped to. 

At each step of the random walk we choose a random neighbor of the current node $v_j$ (see RandomNeighbor($v_j$) in Alg.\ref{alg:RHCF})  with probability ${p_{ij}}$ inversely proportional to the cost $c_{ij}$ associated to the edge ${e_{ij}}$
% The RandomNeighbor($v_j$) routine, in Alg.\ref{alg:RHCF}, selects the next vertex to follow: if at the \emph{j}-th step we are at ${v_j}$, we move to its neighbor ${v_i}$ with probability ${p_{ij}}$. Our random walk runs on the weighted graph: the probability ${p_{ij}}$ of following an edge ${e_{ij}}$, moving from node $j$ to $i$, is given by:

\begin{equation}
\label{transitionprob}
p_{ij} = \frac{ \frac{1}{c_{ij}} A_{ij}}{\sum_k \frac{1}{c_{kj}} A_{kj}}
\end{equation}

where $A_{ij}$ is an element of the adjacency matrix $A$.

The transition matrix $P$ of the graph $G$ is defined as the $N \times N$ matrix where each of its elements is defined as in Eq.\ref{transitionprob}.

Each time we leave a node, we mark it as visited (by removing it from the local copy of the graph $G_p$) and do not allow the algorithm to walk through it again in the current run of the random walk.  
The walk stops when the goal node is found or when we reach a node with all neighbors marked as visited. Each time a path \textbf{P} is generated, we compare it to the ones already found. If the same path was not generated before and it is a valid path, we save it in our homotopy classes set $H$. A path is valid if its last node is $v_g$.  All the visited nodes are then marked unvisited.%: this operation allows the algorithm to find the complete set of paths.

After the $K$ paths have been found, the robot chooses the best one to follow in terms of the social cost function. In the case that the followed path is invalidated by an unexpected obstacle, the planner selects the best path from the set of available paths, given the current status of the robot and of the environment.
\begin{algorithm}
\caption{Randomized Homotopy Classes Finder}          % give the algorithm a caption
\label{alg:RHCF}  
\begin{algorithmic}
  \STATE \bf{function} RHCF($v_s,\, v_g, G,K $)
  \STATE $k=0$

  \WHILE{$k < K$}
 \STATE $\textbf{P} \leftarrow RandomWalk(v_s,\, v_g, G)$
 
 \IF{$\textbf{P} \not \in H$ and $isvalid(\textbf{P})$}
	 \STATE{$H \leftarrow H \cup \textbf{P}$}
     \STATE $k=k+1$
 \ENDIF
 \ENDWHILE

\end{algorithmic}
\end{algorithm}

\begin{algorithm}
\caption{Random Walk}        
\label{alg:RandomWalk}  
\begin{algorithmic}
  \STATE \bf{function} RandomWalk($v_s,\, v_g, G$)
  \STATE $v_j \leftarrow v_s$
  \STATE $G_p \leftarrow G$
  \STATE $\textbf{P} \leftarrow v_j$

  \WHILE{$v_j \neq v_g$}
	\STATE $v_i \leftarrow RandomNeighbor(v_j)$ 
    \STATE $\textbf{P} \leftarrow  \textbf{P} \cup v_i$
    \STATE $G_p \leftarrow  G_p \setminus E(v_j)$
  	\STATE $v_j \leftarrow v_i$

 \ENDWHILE

\RETURN $\textbf{P}$
\end{algorithmic}
\end{algorithm}

\subsection{Cost definition}
\label{subsec:costdef}
Several state of the art approaches focus on finding the $K$-shortest paths \cite{bhattacharya2010search,kuderer2014online, vela2010,vossICRA2015}. Here we are interested in finding the $K$ best homotopy classes for socially-aware motion planning, therefore we compute the edges weights $c_{ij}$ by using a cost function derived from the social force model introduced by Helbing \cite{helbing1995social}.

The cost $c_{ij}$ is defined by the line integral of the pedestrians' interaction forces on the planar curve $s_{ij}$ described by the edge $e_{ij}$ and the edge length $l_s$. 

\begin{equation}
c_{ij} = \int_{s_{ij}} F_s ~ ds + l_s
\end{equation}

The force  $F_s$  represents the force generated from the interactions of all the pedestrians $p_i$ with the robot, defined as pedestrian $p_j$,

\begin{equation}
  \label{eq:interaction-force}
  F_{s} = \sum_{i \in \Persons}  \mathbf{f}_{i,j} 
\end{equation}
with $\Persons = \{p_i\}_{i=1}^{N_p}$ being the set of $N_p$ pedestrians. The forces $\mathbf{f}_{i,j} $ decrease proportional
to the distance of their sources to the robot, and are modelled as
\begin{equation}
 \mathbf{f}_{i,j} = a_{j} e^{\left(\frac{r_{i,j} - d_{i,j}}{b_{j}} \right)} \fdirection_{i,j} 
\end{equation}
where $a_{j}$ specifies the magnitude and $b_{j}$ the range of
the force. The distance $d_{i,j}$ is given by the Euclidean distance between the pedestrian $p_i$ and robot, $r_{i,j}$ is the sum of their radii (we approximate each pedestrian with a circle). The term $\fdirection_{i,j}$ is the normalized vector pointing from $p_i$ to the robot which describes the direction of the force.

\begin{figure}[b!]
  \centering
  \includegraphics[width=0.95\columnwidth]{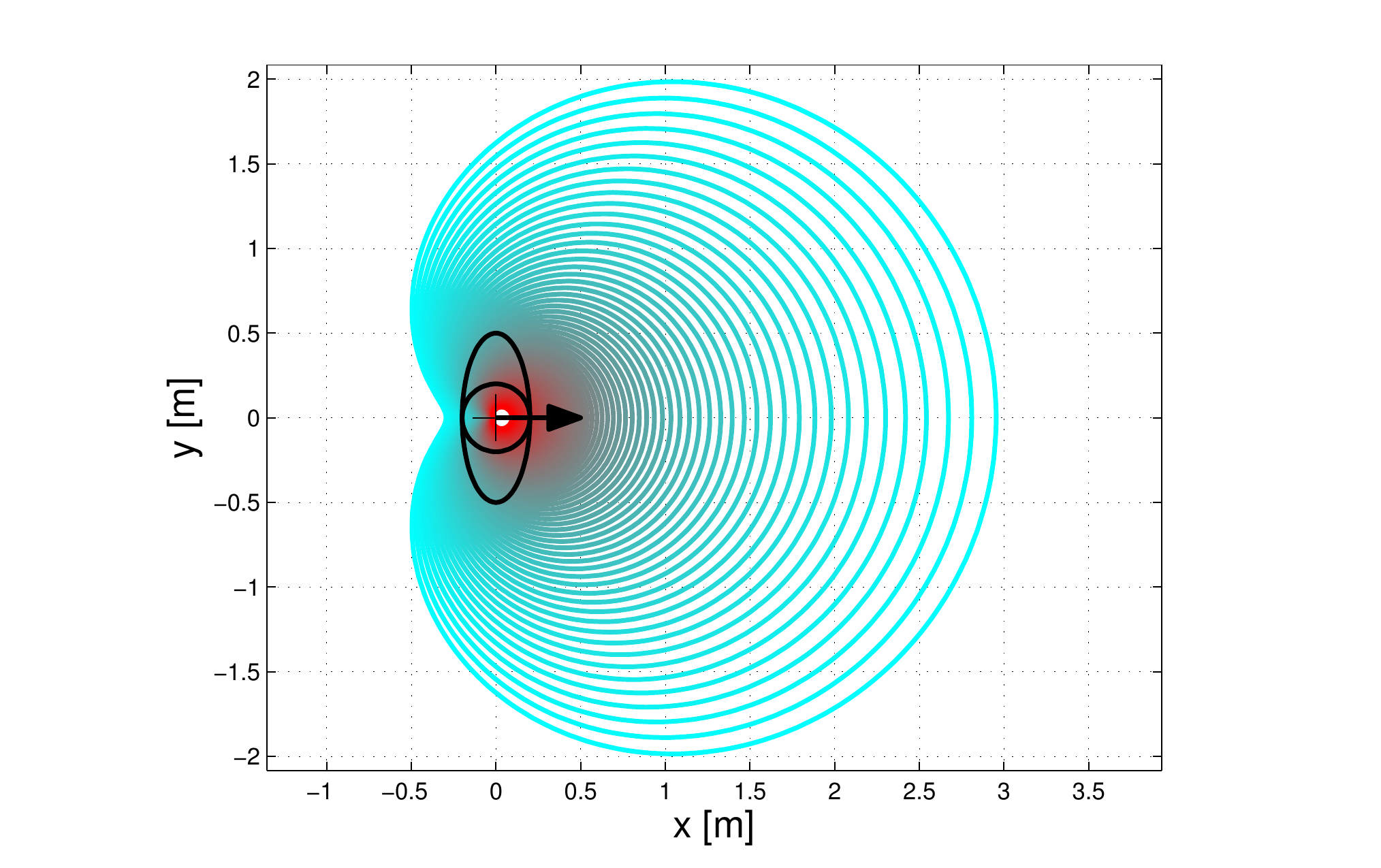}
  \caption{Anisotropic influence model for $\lambda = 0.1$, $a_j = 2$, $b_j = 1$, $r_{ij} = 0.4$ generated by the social force $F_{s}$ for a single pedestrian. \textbf{Red} regions denote higher cost regions.}
  \label{fig:anisoforcemodel}
\end{figure}

To better describe the limited field of view of the pedestrian, the forces are scaled with an anisotropic factor (see Fig.\ref{fig:anisoforcemodel})
\begin{equation}
 \mathbf{f}_{i,j} = a_{j}e^{\left(\!\frac{r_{i,j} - d_{i,j}}{b_{j}} \!\right)}\!\fdirection_{i,j}\!\left(\!\lambda+(1-\lambda)\frac{1 + \cos \left( \varphi_{i,j}\! \right)}{2}\right)
 \label{eq:anisotropic}
\end{equation}
where $\lambda$ defines the strength of the anisotropic factor and
\begin{equation}
 \cos \left( \varphi_{i,j} \right) = -\!\fdirection_{i,j} \cdot \idirection_i .
\end{equation}
with $\idirection_i$ representing the direction of the pedestrian $p_i$.

 Notice that our approach is not limited to a single definition of cost, different definitions from the social navigation literature could be used \cite{kuderer2014online,vasquezIROS14, kruse2014evaluating}.

\section{Experiments}
\label{sec:experiments}

%TODO 
To evaluate our approach in terms of planning performance and quality of the solutions, we design a set of experiments by choosing proper environments and metrics. We compare our approach to Yen's algorithm. In both algorithms we adopt the cost defined in Sec.\ref{subsec:costdef}. The algorithms are implemented in C++. All experiments were running on an ordinary PC with 2.3 GHz Intel Core i5 and 8 GB of RAM. %TODO check Andrey PC details

\subsection{Yen's Algorithm}
In \cite{yen1971finding} Yen introduces an algorithm to find $K$-shortest loopless paths for a given pair of start and goal poses. The algorithm computational upper bound increases linearly with the value of $K$: with modern data structure it can be implemented in $O(KN(M+Nlog(N)))$ worst-case time. We use the C++ implementation introduced by Martins and Pascoal in \cite{martins2003new}, which is reported to have better performance than the straightforward implementation. We compare our approach to the Yen's algorithm, because it finds a set of $k$ best paths with an higher diversity than the ones found by Eppstein's and it was shown by Brander and Sinclair \cite{brander1996comparative} that for small size graphs and paths of small length (like the graphs generated from a Voronoi diagram), it is faster than Katoh's.

\subsection{Environments}
\begin{figure*}[!]
  \centering
\includegraphics[width=0.90\columnwidth]{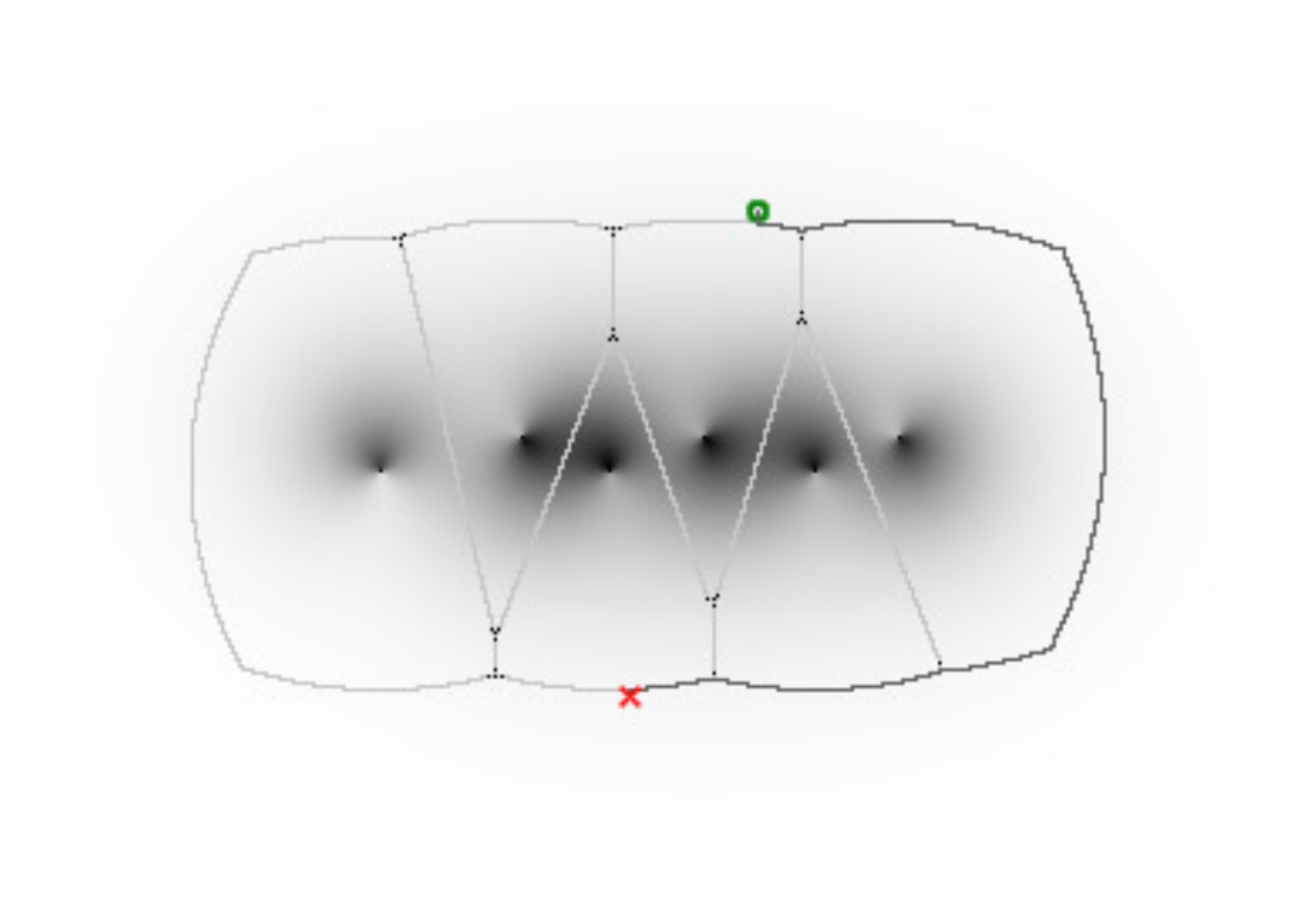}
\includegraphics[width=0.90\columnwidth]{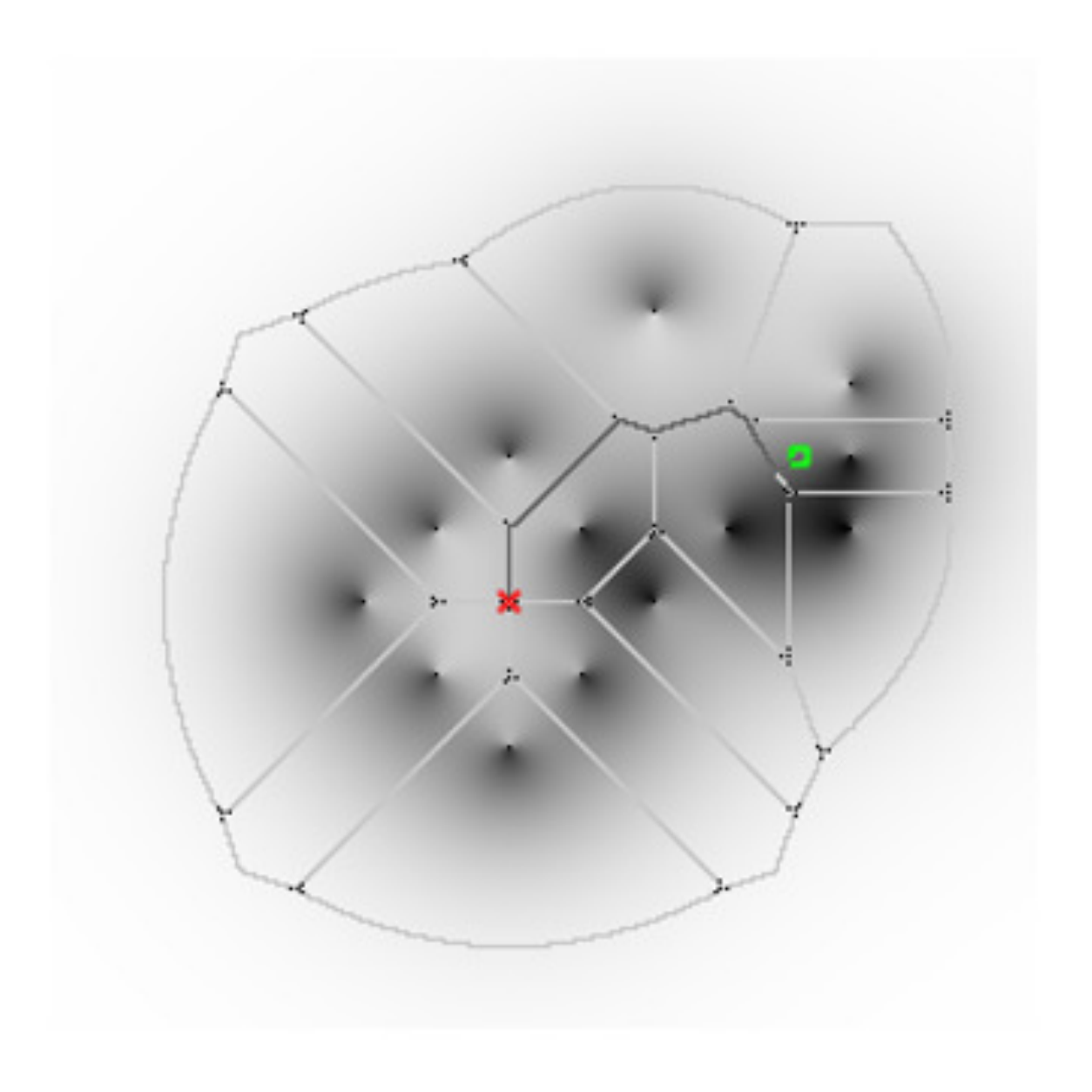}
    \\
  \includegraphics[width=0.90\columnwidth]{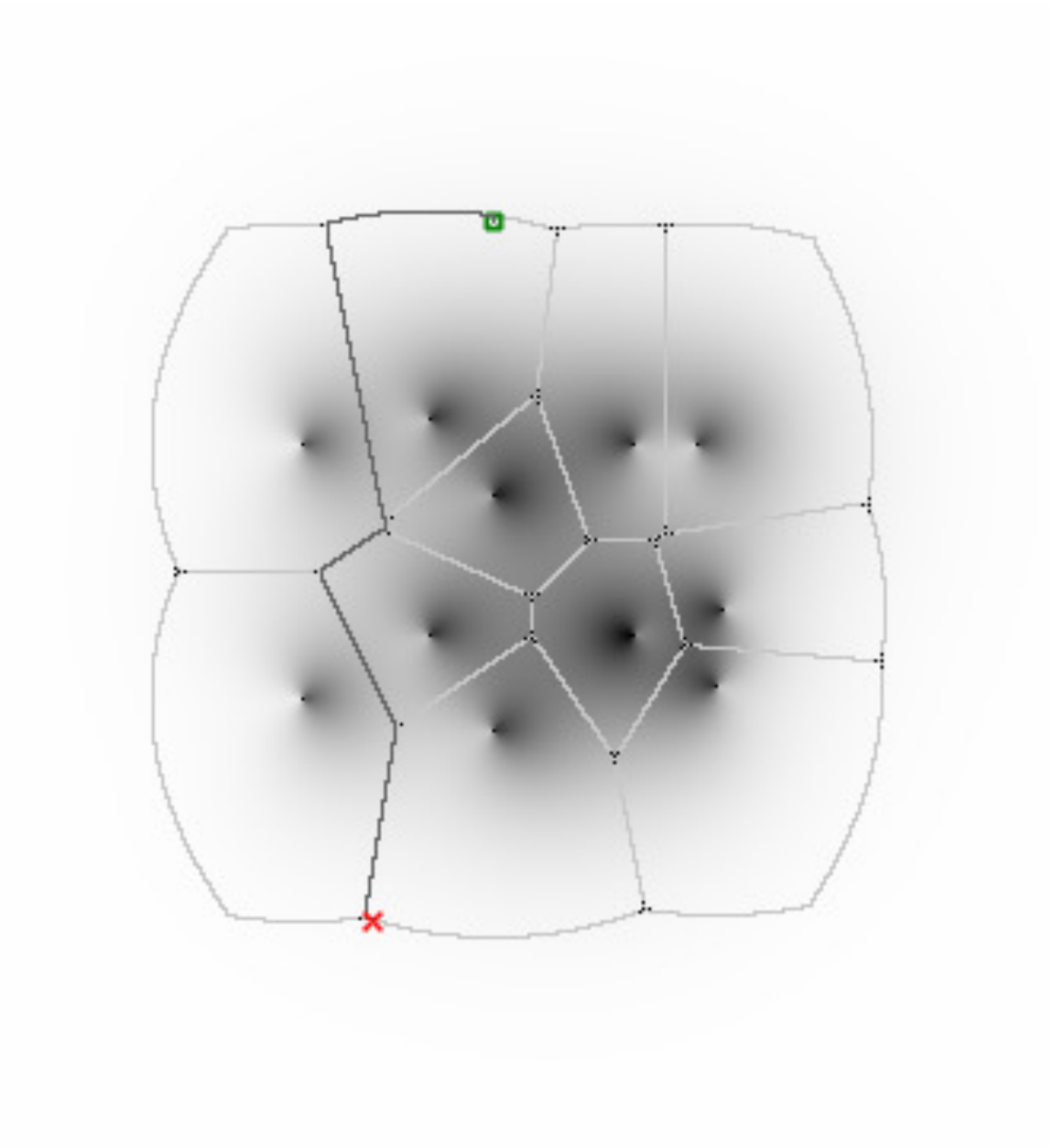}
  \includegraphics[width=0.90\columnwidth]{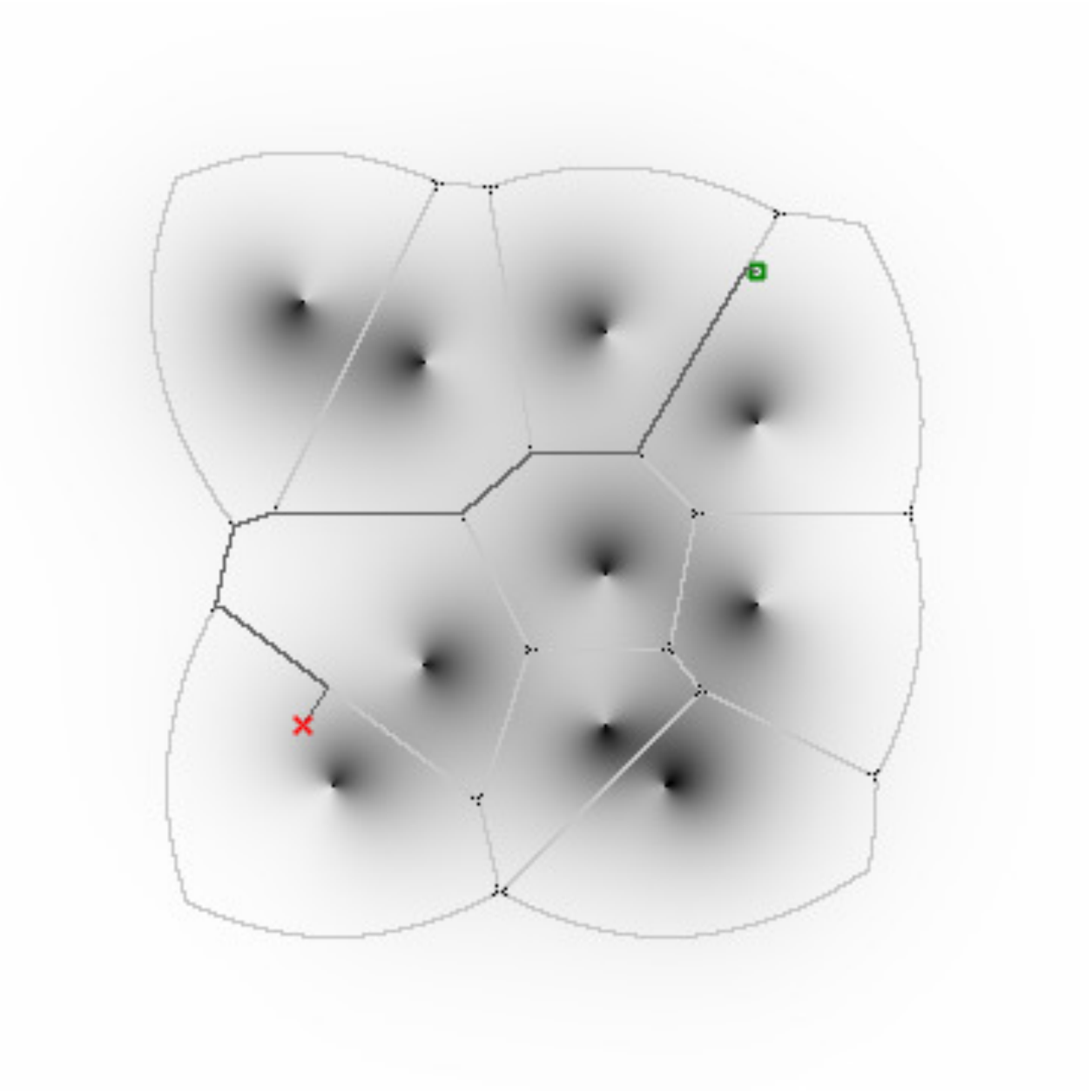}
  \caption{In all the environments, the social force is displayed in grey scale: darker regions have a higher social cost. The peaks of the social force field represent the agent positions. The scenario at \textbf{top left} is the \emph{wall of people} environment. The one at \textbf{top right} part of the figure is the \emph{surrounded} environment. At the \textbf{bottom left} and at the \textbf{bottom right} are respectively the \emph{crowd A} and the \emph{crowd B} scenarios. \textbf{Red} crosses represent the robot position, the goal is displayed by \textbf{green} circles. In all the figures, the edges of the Voronoi diagram are in \textbf{dark grey} and example paths generated by our approach are displayed with \textbf{black} edges.}
  \label{fig:environments} 
\end{figure*}

We design four different environments (shown in Fig \ref{fig:environments}), to stress different properties of the planner and to study how the 
 algorithm behaves in environments of varying complexity. We choose scenarios resembling
potential situations that could occur while a robot is navigating into an airport. In the \emph{wall of people} scenario, the robot needs to find different ways to the goal through a queue of standing people, this scenario has 33 possible homotopy classes. In the \emph{crowd A} (670 homotopy classes) and in the \emph{crowd B} (576 homotopy classes) scenarios, the people are placed in a sparser way forming different groups. In the scenario \emph{surrounded}, that has 1826 possible homotopy classes, the robot is placed in the crowd, surrounded by several people. %The \emph{two people} scenario (with 4 homotopy classes) shows a simple situation where only two people obstruct the way to the goal.

\subsection{Metrics}
To quantify planning performance and difference in quality with respect to Yen's search method we compute the averages of the following metrics:  $T_k$ time to get $K$ different homotopy classes, $nCG_k$ normalized cumulative gain, $RD_k$ robust diversity of a set $\mathbf{P}_K$ of K paths returned by the algorithms. The \emph{robust diversity} measures how large are the intra-set distances between pairs of paths in the set $\mathbf{P}_K$. Let us consider $d_f(p_a, p_b)$ to be the Fr\'echet distance between two paths $p_a$ and $p_b$ evaluated at the vertices, as in \cite{vossICRA2015}. We define $RD_k$ as
\begin{equation*}
\label{equ:robustdiversity}
RD_k = \frac{1}{|\mathbf{P}_K|} \sum_{p_a \in P} \min\limits_{p_b \in P, p_a \neq p_b} d_f(p_a, p_b).
\end{equation*}

The normalized cumulative gain $nCG_k$ is often used to measure the goodness of the ranking results returned by a web search engine algorithm. It computes how far is the candidate ranking set from the ideal ranking set. It is based on the definition of relevance ($rel$) of a single path. In our case the relevance is defined as the inverse of the path cost.
\begin{equation*}
\label{equ:cumulativegain}
CG_k = {\sum_k^K rel_k} 
\end{equation*}
\begin{equation*}
\label{equ:normalizedcumulativegain}
nCG_k = \frac{CG_k}{\max(CG_k) } 
\end{equation*}
To paths with smaller costs correspond higher values of cumulative gain. $nCG_k$ is normalized by the maximum cumulative gain of $k$ best paths generated by Yen's.
   \begin{figure}[!]
  \centering
  \includegraphics[width=0.990\columnwidth, 
  height=0.650\columnwidth]{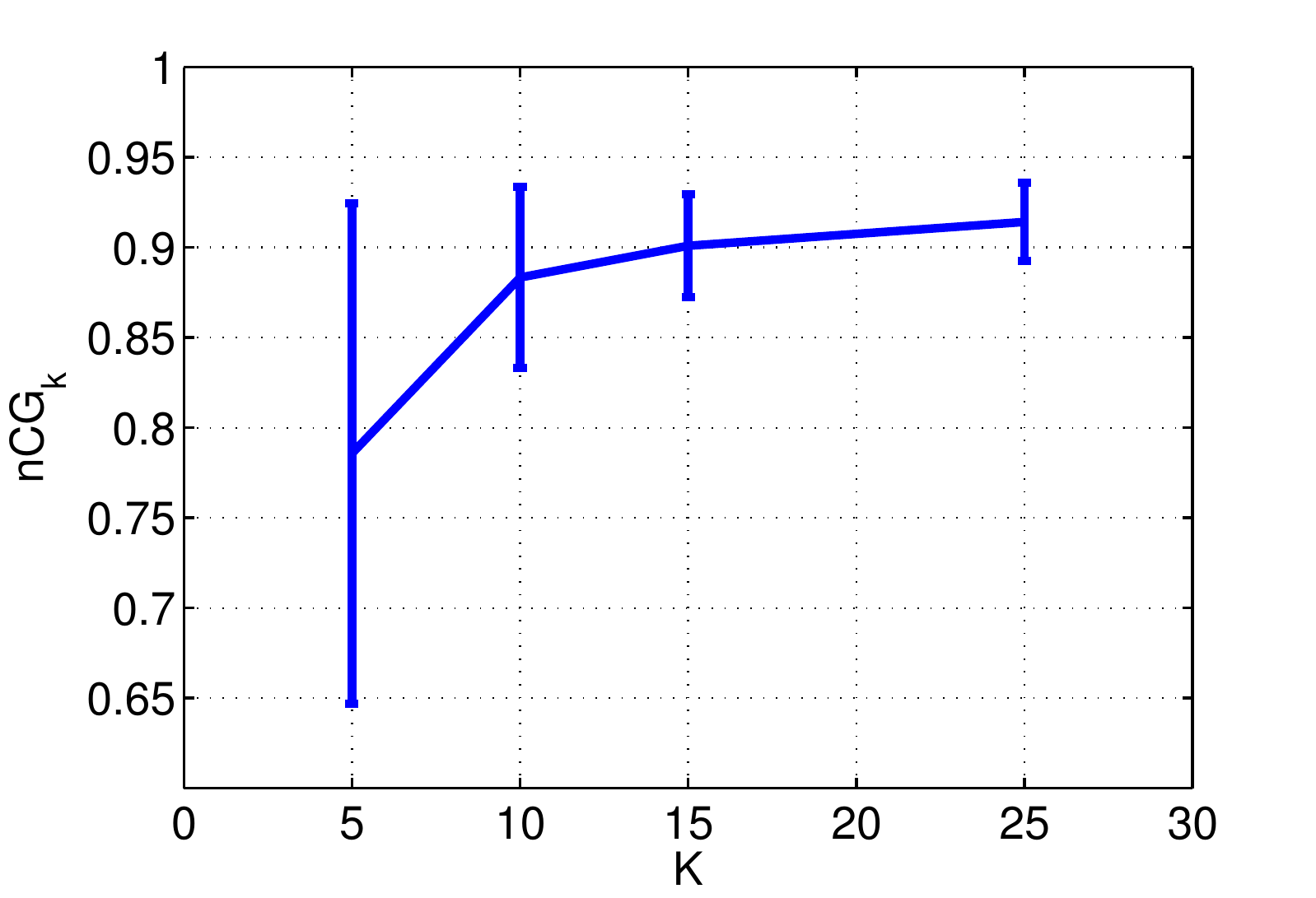}
  \caption{Normalized Cumulative Gain $\mathbf{nCG}_k$ obtained by varying $K$ in the \emph{crowd A} scenario. As $K$ increases, our approach converges to the optimal value of the normalized cumulative gain $\mathbf{nCG}_k$.}
  \label{fig:crowdAcg}
\end{figure}
\begin{figure}[!]
  \centering
  \includegraphics[width=0.990\columnwidth, 
  height=0.650\columnwidth]{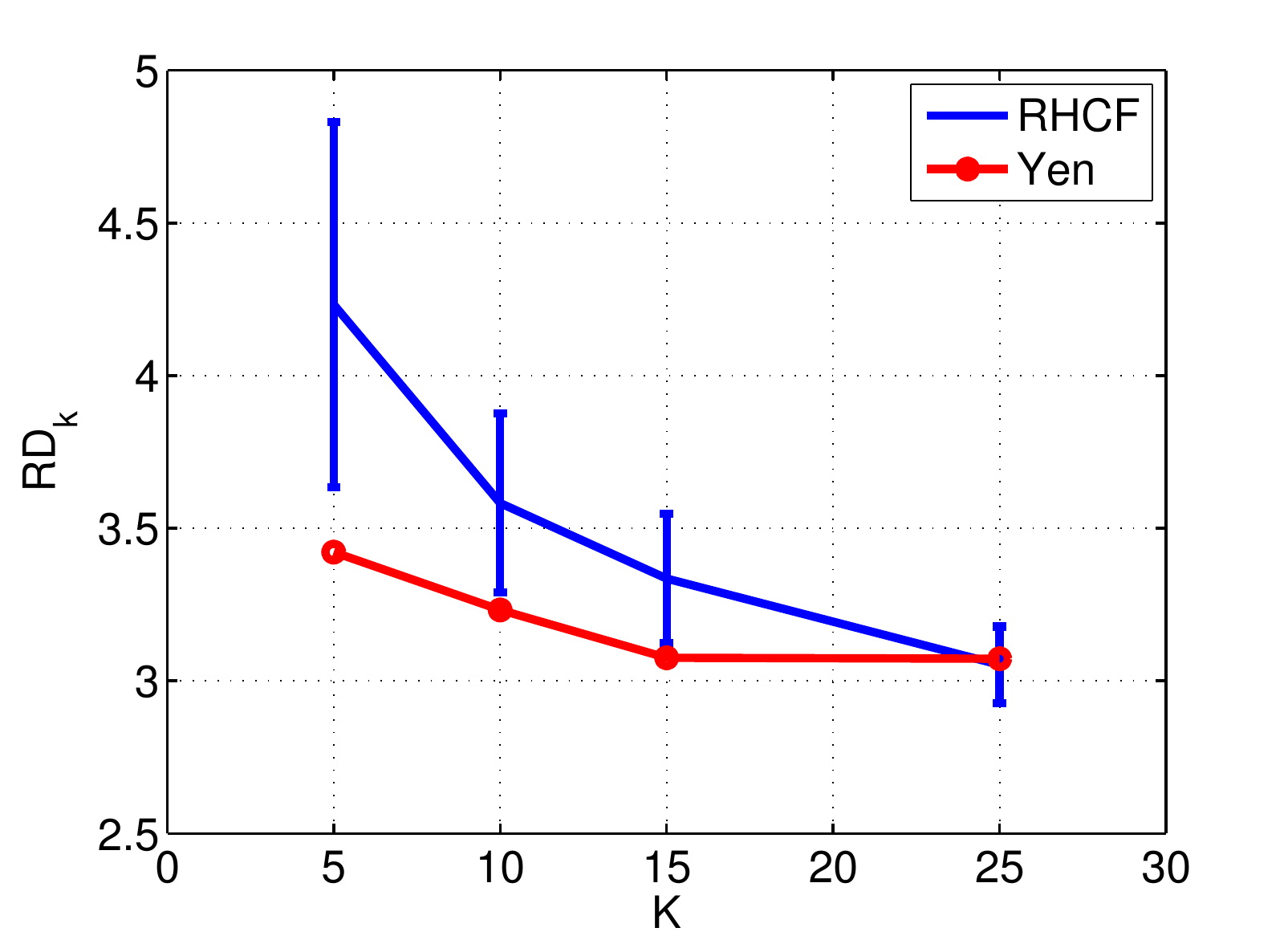}
  \caption{Robust diversity $\mathbf{RD}_k$ obtained by varying $K$ in the \emph{crowd A} scenario. The paths produced by our approach are more diverse than the one generated by Yen's for small values of $K$.}
  \label{fig:crowdADiversity}
\end{figure}
\begin{figure}[!]
  \centering
  \includegraphics[width=0.990\columnwidth, 
  height=0.650\columnwidth]{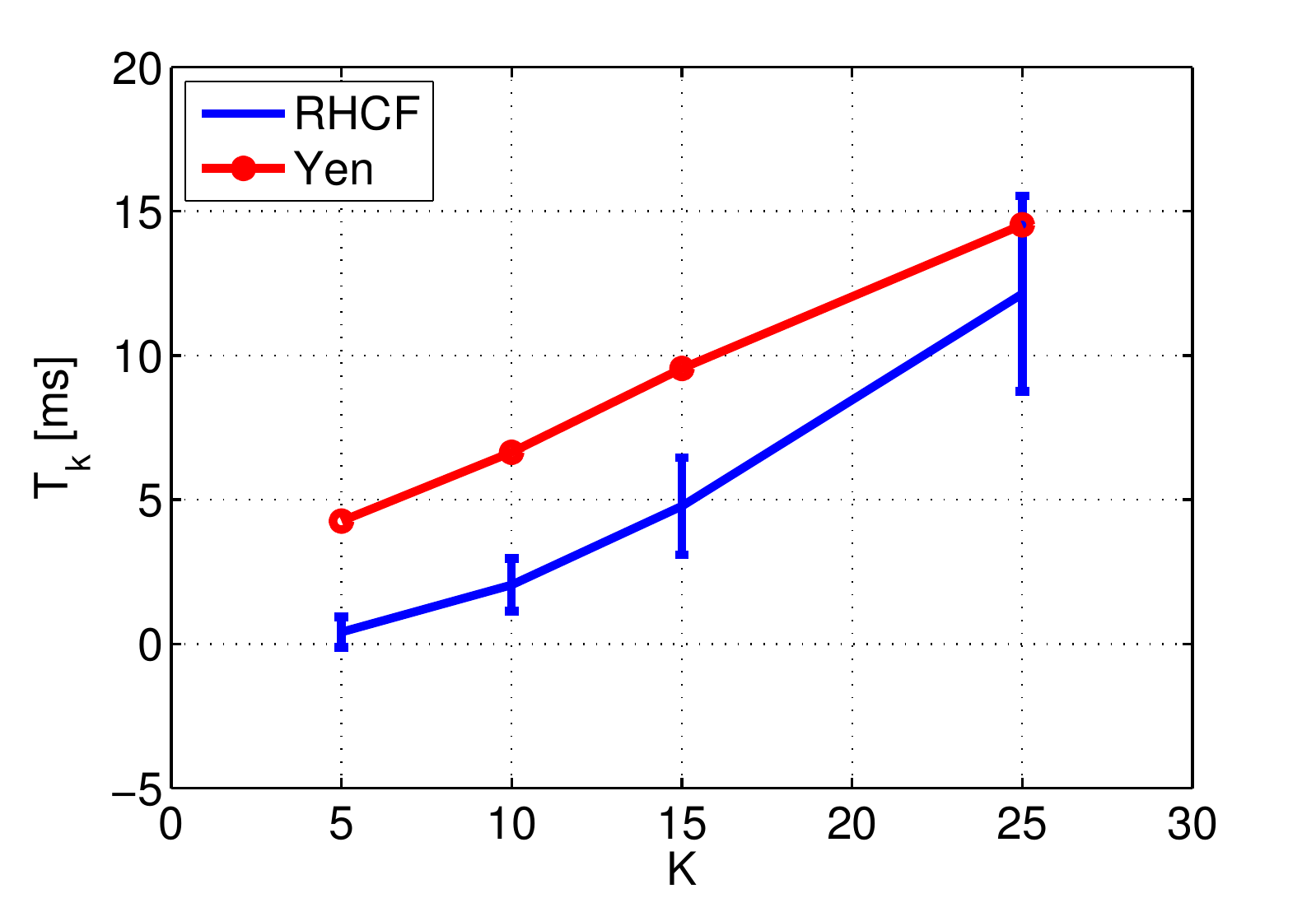}
  \caption{Planning time $\mathbf{T}_k$ obtained with different values for $K$ in the \emph{crowd A} scenario. For small values of $K$, our approach is faster than Yen's algorithm in very complex scenarios (with hundreds of different homotopy classes).}
  \label{fig:crowdAPlanningTime}
\end{figure}
\subsection{Parameters}
In the algorithm only one parameter need to be set: the number of homotopy classes to find. For Yen and RHCF algorithms we find the first 5 paths or homotopy classes (K=5). The parameters of the cost function after several informal validations have been set to (see Fig.\ref{fig:anisoforcemodel}): magnitude of social force to pedestrians $a_j$ to 2, range of social force to pedestrians $b_j$ to 1, strength of the anisotropic factor $\lambda$ to 0.1, $r_{ij}$ equal to 0.4 $m$. 
\section{Results and Discussion}
\label{sec:results}
 Table \ref{Table:DiversityResultsP}, Table \ref{Table:ComulativeGainResultsP} and Table \ref{Table:TimeResultsP} collect the results generated for all the described scenarios, considering the number of homotopy classes to find equal to five ($K=5$).
 Table \ref{Table:TimeResultsP} shows the planning time results: our approach is at least five times faster than Yen's approach.
   Table \ref{Table:ComulativeGainResultsP} details the results related to the cumulative gain, \emph{RHCF} while being faster than Yen's it also finds homotopy classes with a gain close the optimal one (i.e 1): the costs of the solutions generated by the two approaches are only slightly different, the solutions' quality of the two approaches are very similar. 
    Table \ref{Table:DiversityResultsP} details the diversity of the paths generated by both approaches: our method outperforms Yen's in three environments of the four. Only in one scenario, \emph{wall of people}, Yen's finds more diverse paths.
    
   Fig.\ref{fig:crowdAcg}-\ref{fig:crowdAPlanningTime} show the metrics trends for different values of $K$ in the \emph{crowd A} scenario (the same trends are visible in the other scenarios too). For small values of $K$, our approach is faster than Yen's algorithm in very complex scenarios (with hundreds of different homotopy classes), as it is showed in Fig.\ref{fig:crowdAPlanningTime}. In average, when $K$ is greater than one fourth of the total possible homotopy classes in the scenario, the algorithm is slower than Yen's. Our approach has a better robust diversity ${RD}_k$, considering $K$ up to 25 see Fig.\ref{fig:crowdADiversity}: the paths produced are more diverse than the one generated by Yen's for small values of $K$. With a higher value for $K$, see Fig.\ref{fig:crowdAcg}, our approach converges to the optimal value of the normalized cumulative gain ${nCG}_k$ values, the one associated to Yen's rankings. 

\begin{table}[!]
\center
\footnotesize{
\begin{tabular}{ | p{2.1cm} | p{1.9cm}| p{1.9cm}|}
\hline
\multicolumn{3}{|c|}{$\mathbf{RD}_k$} \\
\hline  
\textbf{Scenarios} & RHCF & Yen \\ \hline

Crowd A & $\mathbf{3.76813}$  & $ 2.79823 $ \\ \hline

Crowd B & $\mathbf{4.23284}$ & $  3.4217$ \\ \hline

Wall of People & ${3.7924}$  & $\mathbf{5.19881}$ \\ \hline 
   
Surrounded & $\mathbf{3.76813}$  & $ 2.79823 $\\ \hline 

\end{tabular}
}
\caption{Robust Diversity Results}
\label{Table:DiversityResultsP}
\end{table}

\begin{table}[!]
\center
\footnotesize{
\begin{tabular}{ | p{2.1cm} | p{1.9cm}|}
\hline
\multicolumn{2}{|c|}{$\mathbf{nCG_k}$} \\
\hline  
\textbf{Scenarios} & RHCF \\ \hline

Crowd A & $0.7857$ \\ \hline

Crowd B & $0.7142$  \\ \hline

Wall of People & $0.771$ \\ \hline 
   
Surrounded & $0.7461$  \\ \hline 

\end{tabular}
}
\caption{Cumulative Gain Results}
\label{Table:ComulativeGainResultsP}
\end{table}

\begin{table}[!]
\center
\footnotesize{
\begin{tabular}{ | p{2.1cm} | p{1.9cm}| p{1.9cm}|}
\hline
\multicolumn{3}{|c|}{$\mathbf{T_K}~[ms]$} \\
\hline  
\textbf{Scenarios} & RHCF & Yen \\ \hline

Crowd A & $\mathbf{0.41}$  & $ {4.26} $ \\ \hline

Crowd B & $\mathbf{1.53}$ & $ {4.42}$ \\ \hline

Wall of People & $\mathbf{0.035}$  & $ 2.07  $\\ \hline 
   
Surrounded & $\mathbf{1.05}$  & $ 5.61 $\\ \hline 

\end{tabular}
}
\caption{Planning Time Results}
\label{Table:TimeResultsP}
\end{table}
% ================================================================== %
\section{Conclusions}
\label{sec:conclusions}

In this paper, we introduce the Randomized Homotopy Classes Finder, that finds  homotopy classes in an undirected weighted graph built from a Voronoi diagram. We use the algorithm to find a set of $k$ distinct socially-aware paths from which the robot chooses the best one to follow in terms of a social cost function. 
Our experimental evaluation shows that our approach is faster than Yen's approach. Moreover, as the cumulative gains results show, the paths produced by our approach are not far from the ones generated by Yen's that finds the true $k$ best paths. 
A key property is that our approach computes a set of more diverse paths respect to the baseline: usually different paths share few edges which make them robust to invalidation due to unexpected obstacles.
%We proved that our approach can find all possible paths in a graph in a undirected weighted graph.
%

In future work, we intend to further improve the time performance of the \emph{RHCF} by introducing a discounting factor that biases the search towards a not frequently visited subset of the state space, therefore increasing the probability to generate paths not yet found. %We intend to deeply study the space and time complexity of the algorithm. 
Moreover we are interested to couple our approach with an informed (weighted) Voronoi diagram that implicitly encodes the social context of the scene. Finally, we plan to integrate our algorithm in a hierarchical framework where an optimal sampling-based motion planner generates (locally) optimal kinodynamic trajectories in the best homotopy class found by \emph{RHCF}.
%In future work, we intend to study deeply the space and time complexity of the algorithm and investigate the integration with a low-level planner framework, to generate kinodynamic feasible trajectories. Furthermore given the performances in terms of planning efficiency, we are interested to extend the algorithm to exploit predictions of people and objects in dynamic environments. 
%Furthermore we plan to combine the method with the Inverse Reinforcement Learning framework (IRL): the social cost definition will be replaced with the learned cost maps provided by IRL.
 \balance

% ================================================================== %
\section*{Acknowledgements}
The authors thank Markus Kuderer and Christoph Sprunk for valuable discussions and feedback. This work has partly been supported by the European Commission under contract number FP7-ICT-600877 (SPENCER)
% ------------------------------------------------------------------ %
\bibliographystyle{IEEEtran}
\footnotesize{
\bibliography{palmieriRudenkoIROSWS15}
}
\end{document}